%% file: main.tex
\documentclass[acmtog, authorversion]{acmart}
\acmSubmissionID{431}
\input{defs}
\input{packages}
\acmJournal{TOG}

\copyrightyear{2022}
\acmYear{2022}
\setcopyright{rightsretained}
\acmConference[SA '22 Conference Papers]{SIGGRAPH Asia 2022 Conference Papers}{December 6--9, 2022}{Daegu, Republic of Korea}
\acmBooktitle{SIGGRAPH Asia 2022 Conference Papers (SA '22 Conference Papers), December 6--9, 2022, Daegu, Republic of Korea}\acmDOI{10.1145/3550469.3555411}
\acmISBN{978-1-4503-9470-3/22/12}

\AtBeginDocument{%
  \providecommand\BibTeX{{%
    \normalfont B\kern-0.5em{\scshape i\kern-0.25em b}\kern-0.8em\TeX}}}

\begin{document}

\title{QuestSim: Human Motion Tracking from Sparse Sensors with Simulated Avatars}

\author{Alexander Winkler}
\email{winklera@fb.com}
\affiliation{%
  \institution{Reality Labs Research, Meta}
  \country{USA}
}

\author{Jungdam Won}
\email{jungdam@fb.com}
\affiliation{%
  \institution{Meta AI Research}
  \country{USA}
}

\author{Yuting Ye}
\email{yuting.ye@fb.com}
\affiliation{%
  \institution{Reality Labs Research, Meta}
  \country{USA}
 }


\begin{abstract}
Real-time tracking of human body motion is crucial for interactive and immersive experiences in AR/VR. However, very limited sensor data about the body is available from standalone wearable devices such as HMDs (Head Mounted Devices) or AR glasses. In this work, we present a reinforcement learning framework that takes in sparse signals from an HMD and two controllers, and simulates plausible and physically valid full body motions. Using high quality full body motion as dense supervision during training, a simple policy network can learn to output appropriate torques for the character to balance, walk, and jog, while closely following the input signals. Our results demonstrate surprisingly similar leg motions to ground truth without any observations of the lower body, even when the input is only the 6D transformations of the HMD. We also show that a single policy can be robust to diverse locomotion styles, different body sizes, and novel environments. 
\end{abstract}

\ccsdesc[500]{Computing methodologies~Motion capture}

\keywords{Motion Tracking, Character Animation, Reinforcement Learning, Wearable Devices}

\begin{teaserfigure}
  \includegraphics[width=\textwidth]{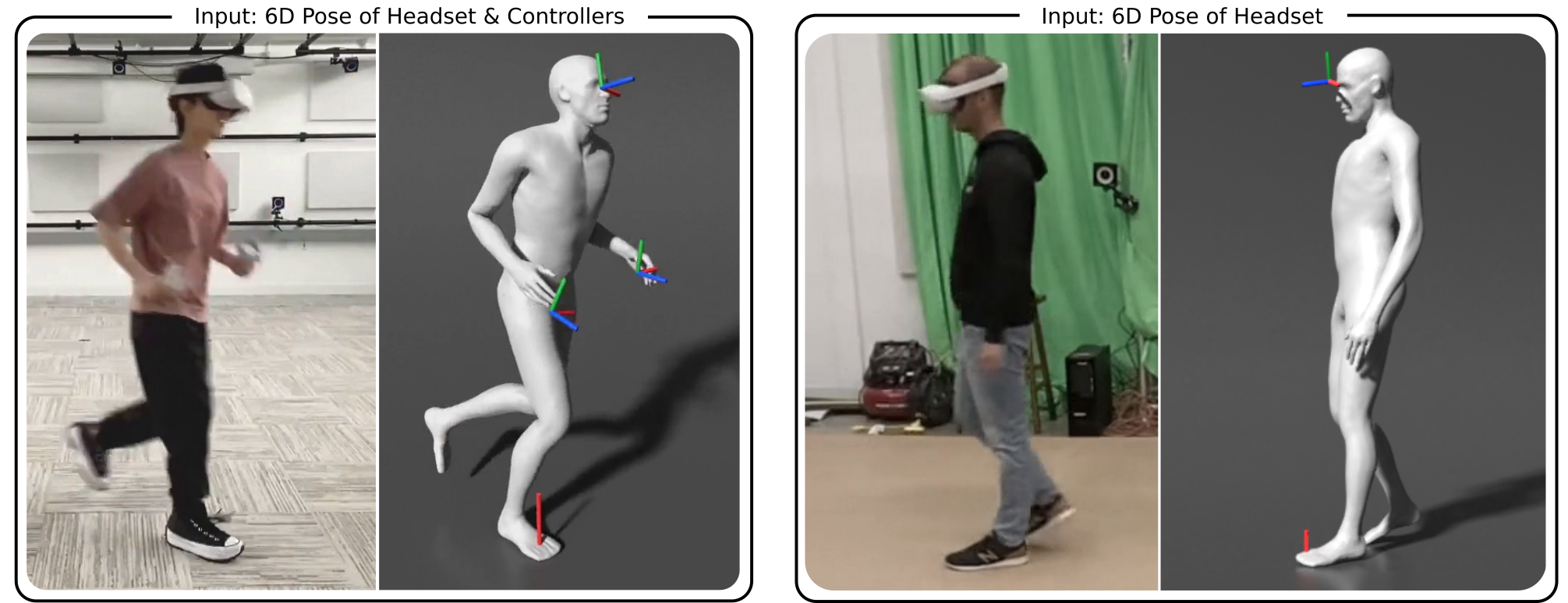}
  \caption{User pose reconstructed from the position and orientation of a headset and two hand controllers (left), or from the headset only (right). The same policy can track users of different sizes (left: 167\,cm, right: 181\,cm). The avatar motion is simulated in a physics engine, which gives access to dynamic quantities such as normal contact forces (red line on the ground). See the \href{https://youtu.be/CkTHsz6Ldas}{\color{blue}video} here.}
  \label{fig:teaser}
\end{teaserfigure}

\maketitle

\input{intro}

\input{related_work}

\input{method}

\input{results}
\input{conclusion}

\newpage
\bibliographystyle{ACM-Reference-Format}
\bibliography{bibliography}

\clearpage
\begin{minipage}{\linewidth} 
\appendix
\input{suppl_material}

\end{minipage}

\end{document}

%% file: defs.tex
\newcommand{\yuting}[1]{}
\newcommand{\jungdam}[1]{}
\newcommand{\jungdamsen}[1]{}

\newcommand{\fref}[1]{Figure~\ref{#1}}

\newcommand{\sref}[1]{Sec.~\ref{#1}}
\newcommand{\tref}[1]{Table~\ref{#1}}

\newcommand{\quest}{Quest~}

\newcommand{\ndof}{33}
\newcommand{\nlinks}{16}



%% file: packages.tex
\usepackage{ dsfont }

\usepackage{subcaption}
\usepackage{soul}

%% file: intro.tex
\section{Introduction}
A promise of AR/VR (Augmented, Virtual Reality) is that it will enable richer forms of self-expression and social experience compared to 2D video. This could be achieved through avatars that accurately capture a user's movement and body language. To enable this, we need sensors and methods to faithfully reproduce the full body motion of a user in real-time.      

Optical \textit{marker-based} solutions~\cite{vicon123} are often used in industries or research labs that require high precision. However, the setup is complex: It requires multiple cameras placed throughout the room, as well as attaching and calibrating markers on the users to be captured.
A solution with less friction is \textit{markerless} motion capture, which does not require any markers attached to the user. However, the sensor still needs to observe the user at all times, so moving between rooms or large scale motion capture is difficult. 
This motivates motion capture from \textit{wearable} sensors, which rely only on sensors attached to the user with no other external sensing modality. One type of wearable sensor are Inertial Measurement Units (IMUs) that can capture both linear and angular motion. Since IMUs are prone to drifting, current HMDs often fuse this acceleration data with camera information (SLAM) to estimate their location \cite{slam:whyte:2006}. This results in reasonable estimates of the global position and orientation of the headset and the controllers. And since the sensors are wearable they can be used across rooms and even outside. 

However, sensor signals that are accessible from AR/VR devices are sparse, with no information about the lower body. To reconstruct full body poses from this data, part of the human pose must be synthesized. Purely kinematic approaches have difficulty synthesizing this missing information in a believable way, especially from sparse inputs, as the space of all possible human poses is vast. This can lead to unnatural artifacts such as jitter, foot skating, and unstable contacts. In this work, we incorporate an off-the-shelf physics simulator into the tracking pipeline in order to constrain the solution space to physically valid poses to mitigate some of these artifacts. 

As contribution we show that sparse upper-body sensors carry enough signal, when combined with physics, to predict the lower-body pose, even from the HMD alone. We demonstrate this by tracking users of different heights from real-world sensor data with a single policy, trained end-to-end with deep reinforcement learning. This creates motions with less artifacts such as foot skating compared to kinematic approaches. The simulated environment can also be used to adapt motions (e.g. to rough terrain), to better fit into the virtual environment.


%% file: related_work.tex
\section{Related work}
We categorize approaches to human motion tracking based on the type of sensor used as input. As the input signal becomes sparser, more of the pose reconstruction must be synthesized. Physics is a sensible prior when reconstructing those aspects of the pose which are not observed by any sensors. Therefore we end this section by reviewing physics-based approaches using Reinforcement Learning (RL), and contrast them to kinematic-based approaches.

\subsection{Vision Sensors}
Generating 3D full body poses based on camera images can be used for motion tracking. This image can be used to predict parameters for a parameterized and differentiable statistical human body model such as SMPL~\cite{SMPL:SIGA:2015, FrankMocap:ICCV:2021,Kanazawa:2019:CVPR, xu2019denserac}. Oftentimes, camera pixels are preprocessed to extract body keypoints \cite{OpenPose:PAMI:2019} or body correspondences from the image \cite{Densepose:CVPR:2018}. One of the difficulties when using monocular cameras is the depth ambiguity, e.g. a short person closer to the camera looks identical to a taller person further away. This depth ambiguity can cause reconstructed 3D poses to accurately match the camera image when viewed from the same angle, whereas different views reveal unnatural leaning, scale and pose. Furthermore, if every frame is reconstructed individually, continuity between frames can be difficult and can cause poses to jitter. Constraining the poses through physics-based priors can help to mitigate these issues \cite{rempe2021humor}.

\subsection{IMU Sensors}
Another popular approach is to use sensors mounted on a user's body, for example inertial measurement units (IMUs). They are small, lightweight and don't exhibit occlusion, since they are based on acceleration signal, versus vision. This allows them to be used across rooms, as well as outdoors and makes them agnostic to light and weather conditions. An offline method using IMUs was proposed by Marcard et al.~\shortcite{Marcard:SIP:2017}, where the system optimizes the pose parameters of the SMPL model so that it matches the sensor's signal. The system performs best when it can access the entire sensor signal trajectory. In order to overcome this offline constraint, deep learning based approaches using mocap data paired with IMU signals were proposed. These models produce a \textit{local} joint angle pose in realtime ~\cite{DIP:SIGA:2018,Nagaraj:APPIS:2020}. 
Methods that estimate the full pose, including global 6D root from IMU signal were explored by \cite{TransPose:SIG:2021, jiang2022transformer}. Since kinematic pose prediction can be jittery or drift, these methods use predicted foot contacts, \emph{Stationary Boundary Points}, or terrain prediction to improve the synthesized poses. Apart from IMUs, sensors based on electromagnetic (EM) fields have also been used to reconstruct full-body poses \cite{kaufmann2021empose}. However, most of the above approaches require sensors attached to the lower body.

\subsection{HMD Sensors}
As Head Mounted Devices (HMDs) for AR/VR are becoming more widely available, methods that generate full-body poses only from HMD and controllers are being explored.  Dittadi et al.~\shortcite{dittadi2021full-body} proposed a framework based on a variational autoencoder (VAE), where a VAE is first learned with all poses in the dataset, then the decoder is combined with another encoder which uses IMU signals as input. Aliakbarian et al.~\shortcite{aliakbarian2022flag} utilized a flow-based model instead of VAEs, where the invertible nature of flow-based models enables learning a shared latent space in a single learning phase. Jiang et al. \shortcite{AvatarPoser2022} uses a Transformer architecture to predict the global skeleton state from only the HMD and controller poses. However, since these approaches are kinematic and don't enforce physical constraints, they can suffer from artifacts such as foot-skating and jitter.









\subsection{Physics-based Approaches}
Many of the previous approaches are \textit{kinematic}, meaning that the model has no notion of masses, inertias or forces as it synthesizes the poses. However, obeying physical laws makes synthesized motions more believable. This physics prior is especially important when the problem is under-constrained, where many solutions fulfill the sensor constraints but only a few are desirable. This is the case when using sparse sensor signal (head and controllers) to reconstruct physically accurate full-body poses. 

Shimada et al.~\shortcite{PhysCap:SIGA:2020} optimized the output kinematic motion predicted from a monocular video where physics laws are used as soft constraints. Yi et al.~\shortcite{PIP:CVPR:2022} added a physics-based optimizer as a refinement process to further improve motions generated by kinematic methods such as \textit{TransPose}~\cite{TransPose:SIG:2021}, showing impressive results for IMU-based pose reconstruction. We show that we can generate motions of comparable quality without sensors on the lower body.  

However, many methods still allow physics to be slightly violated, e.g. they are not enforced as hard-constraints. This can cause the posture of the characters to be abnormally tilted or for generated motions to appear to be floating in the air. 
We build much of our work on RL-based imitation learning, which adds a physics simulator as the final step before the pose generation, thereby enforcing physics as hard-constraints that cannot be violated \cite{2018-TOG-deepMimic}. The controller outputs either target joint angles or torques which are used by the physics simulator to generate the next pose of the simulated character. This stream of research traditionally focused on imitating motions from \textit{dense}, \textit{full-body} user observations. Here, a variety of methods have explored how a single policy can imitate diverse motions, such as walking, jogging, break-dancing etc \cite{won2020scalable,Park:2019,Bergamin:2019:DReCon,Levi:2021:SuperTrack,Chentanez:2018}. This becomes especially relevant for motion tracking, where we cannot know in advance what type of motion the user will perform, but require the policy to be able to track it. Another line of research is training on large mocap datasets to generate fundamental "building blocks" of motions, that can later be reused to achieve other downstream tasks ~\cite{2021-AMP-Peng,Merel:2020:CatchCarry,Won:2021:FairPlay,Peng:2019:MCP}. In the context of motion tracking these insights could be used to track the motion using a specific style. 

Reinforcement Learning has also been used to imitate kinematic motions generated from sparser observations, such as monocular video~\cite{Peng:2018:SFV,Ri:2021:HDfMV}. It has been successfully combined with egocentric video to synthesize poses and object interaction \cite{luo2021dynamics} and predict future motion \cite{yuan2019ego}. Luo et al.~\shortcite{luo2021dynamics} learned a universal controller and a kinematic policy and used it to generate simulated motions from a single-view egocentric video. In contrast to these video-based approaches, we show motion synthesis using only the 6 DoF state of the headset and controllers. We use a simple architecture, consisting of a single MLP trained end-to-end. Since our physics simulation doesn't use additional non-physical forces, our approach generates high-quality and believable motions.


%% file: method.tex
\section{Method}
Traditional RL-imitation frameworks often assume that full-body, noise-free observations of the reference are available, which is not the case when tracking a user from only a HMD. In the following we detail our architecture based on Reinforcement learning, with important modifications for motion-tracking of users from sparse and real-world sensors.

\subsection{RL background}
We use reinforcement learning to train the character. The goal is to find a policy $\pi_\theta$ to maximise the expected discounted return
\begin{equation}
J(\pi_\theta) 
= \mathds{E}_{\tau \sim \pi_\theta}
\left[\sum_0^T \gamma^t r_t\right],
\label{eq:return}
\end{equation}
where $\theta$ represents the weights of a neural network, $\gamma \in [0,1]$ is the discount factor, and $\tau = (s_0, a_0, \dots, s_{T+1})$ represents trajectories collected by the policy interacting with the environment.
The probability for a specific trajectory $\tau$ when using the current policy $\pi_\theta$ is given by $P(\tau|\theta) = P(s_0)\prod_{t=0}^T P(s_{t+1}|s_t, a_t) \pi_\theta(a_t|s_t)$, where $P(s_0)$ is the initial state distribution and $P(s_{t+1}|s_t, a_t)$ represents the dynamics of the physics simulator. This probability is used to calculate the expectation $\mathds{E}_{\tau \sim \pi_\theta}$ which quantifies the expected return. To learn the policy weights $\theta$, we use the proximal policy optimization (PPO) algorithm~\cite{2017-PPO-Schulman}.

\subsection{Overview}
\begin{figure}[ht]
\centering
\includegraphics[width=1.0\linewidth]{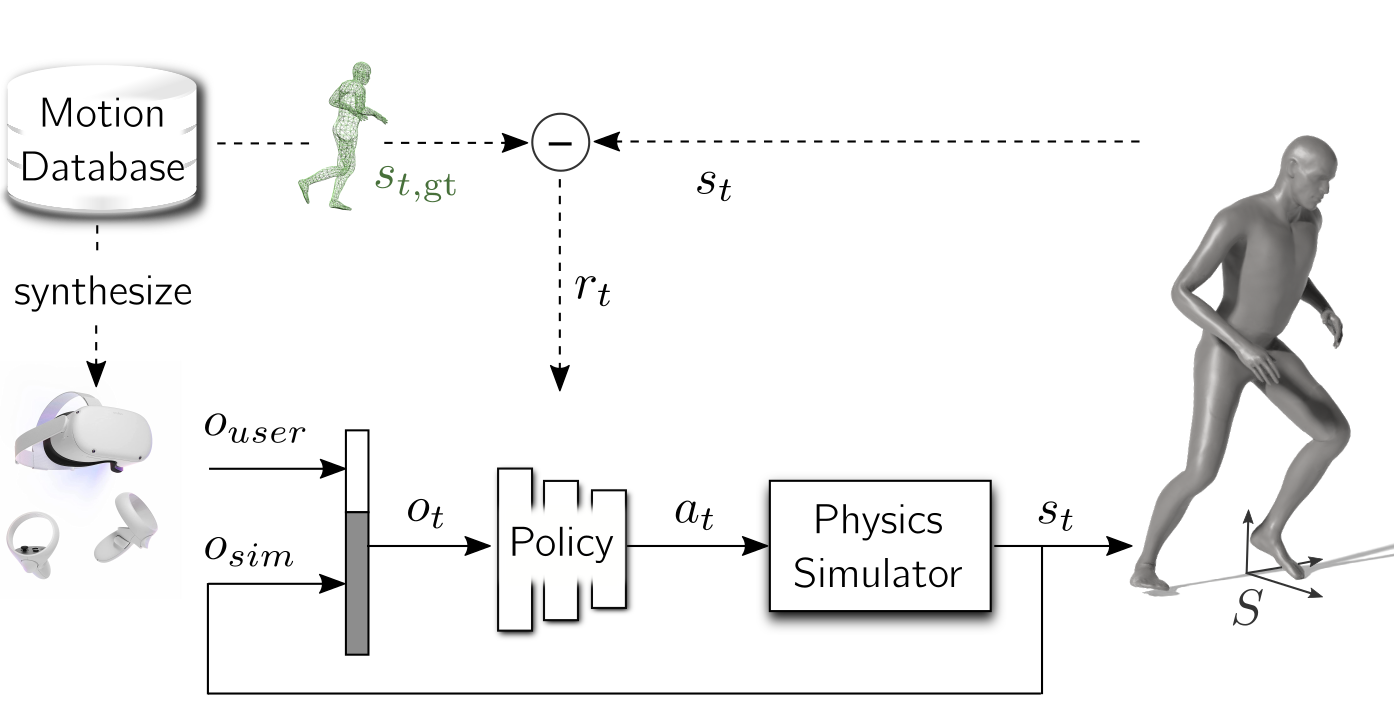}
\caption{Overview of our Imitation Learning architecture to track users from sparse sensor data. The dotted paths are only needed to train the policy network (MLP). During inference, this network produces torques for a physics simulator that generates the pose.}
\label{fig:overview}
\end{figure}

The goal is to reconstruct the full body pose of a user from sparse user observations $o_{\text{user}}$ as shown in \fref{fig:overview}. This output pose $s_t$ is generated by a physics simulator. The simulator is driven by joint torques, which are produced by a neural network termed "Policy". We use Reinforcement Learning, together with an imitation objective~\cite{2018-TOG-deepMimic}, to train the policy to produce torques that track a user. During training we use a scalar reward $r_t$ that captures the goal of producing a pose $s_t$ that is as close as possible to the corresponding ground-truth pose $s_{t,\text{gt}}$ from the motion database. 

We hypothesize this architecture works well with sparse input data because of the high quality full body supervision signal during training, as well as because of the feedback of the simulated state back into the policy. Since the user observations are sparse and don't provide much information about the user, considering the current state of the physically simulated character significantly reduces ambiguity on possible next poses. This allows the policy to use this \textit{internal} state to decide on optimal actions. For kinematic tracking approaches, this feedback loop is conceptually similar to Recurrent Neural Networks (RNNs) or "Pose Priors". However, complex architectures such as RNNs or Transformers are not required in our case and a 3-layer MLP policy is sufficient.

The policy requires observations as input in order to determine the torque at every simulation step. The observations can be split into three sections discussed in the following: the observations coming from the simulated character, the sparse observations coming from the sensors worn by the user, and the scale of the user. 

\subsection{Simulated Character Observations}
\label{sec:sim_obs}
The simulated avatar has \ndof~degrees-of-freedom (DoF). It is fully observable, allowing the policy access to any values that are helpful for determining the torques. We use the joint angles $o_{\text{sim},q} \in \mathbb{R}^{\ndof}$ and joint angle velocities $o_{\text{sim},\dot{q}} \in \mathbb{R}^{\ndof}$. Even though redundant, we also give the policy access to the Cartesian positions and orientations of each link, which speeds up training. All the Cartesian positions $o_{\text{sim},x} \in \mathbb{R}^{\nlinks \times 3}$ are expressed with respect to frame $S$, which is located on the floor below the avatar and rotates according to its heading direction (see \fref{fig:overview}). This allows the policy to learn torque mappings independent of the heading direction. The link orientations $o_{\text{sim},R} \in \mathbb{R}^{\nlinks \times 6}$, also in frame S, are encoded by the first two columns of their rotation matrices. This has been shown to be a favorable orientation representation for neural networks compared to quaternions or angle axis, which can introduce discontinuities. We also observe dynamic quantities such as contact forces of each foot $o_f \in \mathbb{R}^{2 \times 3}$, which allows the policy to reason about contact states when determining the torques. The policy also observes the simulated avatar's linear and angular velocity of each of the links, which is necessary in order to take the inertia of the character into account when producing torques. In total the simulated avatar is observed by $o_{sim} \in \mathbb{R}^{312}$. 

\subsection{Synthetic Training Data}
Training the policy requires sensor data of the user paired with ground-truth poses $s_{t, \text{gt}}$ from which the reward is computed. 
Instead of recording a new dataset capturing the full-body pose of a user (e.g. with a marker-based setup) \textit{while} wearing a headset, we synthetically generate this paired data. We offset the ground-truth head and wrist joints to emulate the position and orientation of a headset and left and right controllers as if the subjects were equipped with the devices. Since the real sensor signal is sufficiently clean, noise was not added to the offsets.

Our in-house Mocap data consists of ~8 hours of motion clips of 172 subjects. Specifically, the dataset contains 130 minutes of walking, 110 minutes of jogging, 80 minutes of casual conversations with gestures, 90 minutes of whiteboard discussion and 70 minutes of balancing. Existing datasets lacked either subject diversity or motion diversity.

\subsection{User Pose Observations}
The sensor data, either coming from the real headset or synthetically generated for training, is given by the position and orientation of the headset $h$, the left controller $l$ and the right controller $r$.
Like the Cartesian observations of the simulated character, all user positions and orientations are relative to frame $S$ as shown in \fref{fig:overview}. The orientations ${}_S R_i \in \mathbb{R}^6$ denote the first two columns of the rotation matrix of the head, left and right controller relative to frame $S$ (see \sref{sec:sim_obs} for details). 
\begin{equation}
o_{\text{user, t}} = [h_S, {}_{S} R_h, \:l_S, {}_{S} R_l, \:r_S, {}_{S} R_r]. 
\end{equation}
The policy also has access to 6 future user observations. This gives the simulated avatar the ability to better anticipate and thereby track a user's motion by knowing what they will do next. The trade-off is that relying on future poses introduces a 160ms latency, which makes real-time tracking less responsive. In total the pose observations coming from the user are $o_{\text{user}} \in \mathbb{R}^{6 \times 3 \times (3+6)} \in \mathbb{R}^{162}$.

\subsection{User Scale observations}
\label{sec:scale_obs}
Many imitation learning approaches generate policies that are specific to one character of a particular scale. In order to deliver a general motion tracking solution, we want to be able to track users of any scale (tall or short). One approach is to generate individual policies per user scale and then blend them during inference. The downside is that this requires a unique policy for every user scale and there might not be sufficient mocap training data available for each user. So instead, we learn a single policy which generalizes to various user scales. 
To achieve this we give the policy access to the user scale in form of a scalar value, which specifies the user's height in meters.
\begin{equation}
    o_{\text{user, scale}} \in \mathbb{R}.
\end{equation}
This allows the policy to learn to adjust torques based on the user, for instance apply larger torques when tracking taller users with larger masses and inertia.

During training, when imitating a clip recorded by one of the 172 individual subjects, we use the first pose in the clip, which is an A-pose, to extract the height of the subject. This single scale value is used to initialize a simulated character with approximately the same scale. During inference, when a user wears the headset, we require the user to initially stand upright and use the height estimated by the headset to initialize their avatar.

\subsection{Reward}
\label{sec:reward}
The reward $r_t$ is used to generate a desired behavior of the character. The goal for the simulated character is to imitate a user's motion as close as possible. We build on the imitation reward introduced by Peng et al. \shortcite{2018-TOG-deepMimic}. Since during training the sparse observations are synthetically generated from the full-body mocap pose, the corresponding full-body pose $s_{t,\text{gt}}$ is known and can be used to formulate this reward. This way the policy has dense supervision during training while requiring only sparse data during inference. Our reward function is given by the dot product
\begin{equation}
r_t = \mathbf{w} [r(q), \:r(\dot{q}), \:r(x), \:r(\dot{x}), \:r_f]^T, 
\end{equation}
where the terms $r(q)$ and $r(\dot{q})$ quantify the difference of joint angles and joint angle velocities between the simulated avatar and the ground truth mocap pose and $r(x)$ and $r(\dot{x})$ quantify the difference in Cartesian positions and velocities of each joint. Each term is expressed using a Gaussian kernel
$
r(s) = \exp({-k_{s}\sum_j \lVert {s}_{sim} - {s}_{\text{gt}} \rVert_2^2 }),
$
where ${s}_{sim}$ and ${s}_{\text{gt}}$ are the particular representation of the simulated character and the ground truth pose respectively and $k_s$ is the sensitivity of the kernel for each representation. Weights $\mathbf{w}$ and kernel sizes $\mathbf{k}$ can be found in the Appendix.

Using only the above terms when no lower-body user observations are available results in the simulated avatar taking short, high-frequency steps, which look unnatural. The additional reward term 
$
r_f = \exp(k_f\sum_{i=L,R} \max(0, f_{y,i,{\text{prev}}} - f_{y,i}))
$
reduces this, by penalizing an abrupt decrease in vertical contact force $f_y \in \mathbb{R}$, where $L$ and $R$ denote the left and right foot, and $k_f$ the sensitivity of the Gaussian kernel. This encourages the avatar to first naturally unload a leg before lifting it, but still allows to forcefully step down to make contact.

\subsection{Training}
\label{subsec:training}
The policy directly outputs torque values instead of PD target angles, which tracked stably and reduced code complexity compared to e.g. \textit{Stable PD} \cite{2011-Tan-SPD} or other more involved controller formulations. 
We run the policy and simulation at 36 frames per second (fps), which we found was the largest stable timestep. This also facilitated downsampling of the HMD sensor data during inference, which is provided at 72 fps. As a physics simulator we use Nvidia's \textit{PhysX}, which has been wrapped by the RL training framework \textit{IsaacGym} \cite{2021-IsaacGym}. This allows us to simulate physics on the GPU and gives access to the simulation data through PyTorch tensors \cite{paszke:2019:pytorch}. We combine \textit{IsaacGym} with the open-source, GPU-ready PPO implementation \textit{rsl\_rl} \cite{2021-Nikita-Isaac-Code, 2021-Nikita-Isaac-Paper}. This allows us to train 4000 characters in parallel on an NVIDIA GTX 3080, which trains to 14 billion policy steps, equivalent to 13 years of human experience, in 48 hours. This is roughly 2 orders of magnitude faster than CPU-based approaches. For a detailed description of RL algorithm parameters see Appendix.


%% file: results.tex
\section{Results}
\begin{figure*}[tbh]
\centering
\includegraphics[width=\linewidth]{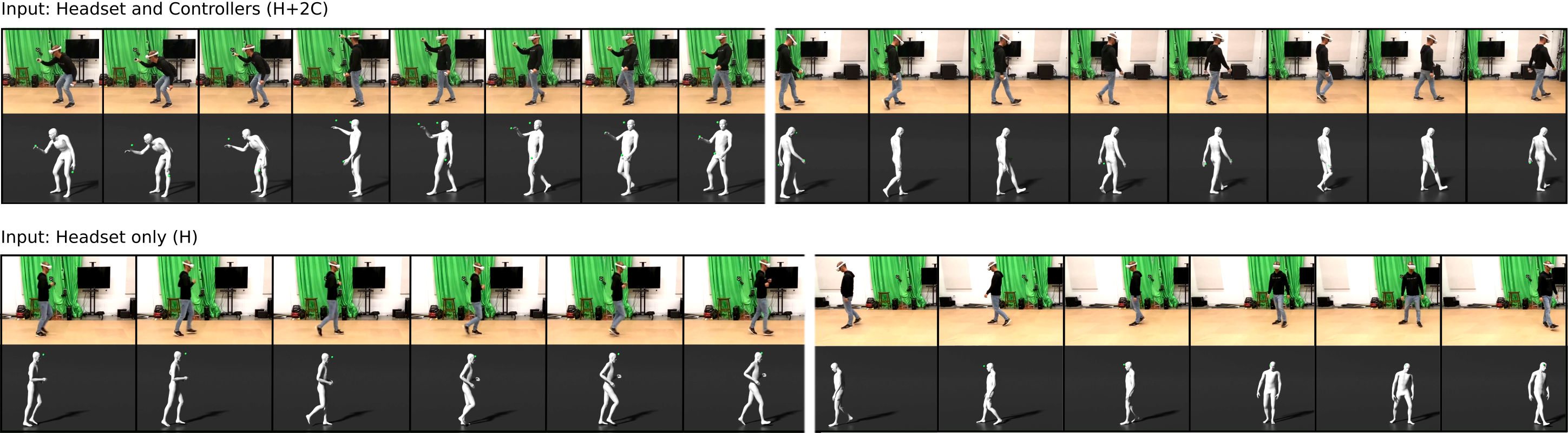}
\caption{Qualitative evaluation of pose reconstruction on motion sequences not used during training, corrected for latency. Most of the reconstructed poses match the video reference footage, despite having no lower-body sensor signal available for reconstruction. The four sequences demonstrate writing on a whiteboard, walking, jogging, and backwards walking with turns. Readers are encouraged to view the \href{https://youtu.be/CkTHsz6Ldas}{\color{blue}video} for more examples.}
\label{fig:pt_new}
\end{figure*}

We show that the positions and orientations of Meta's \quest headset and controllers contain enough signal to reasonable estimate the full-body pose of a user. The framework is able to distinguish between various locomotion modes, turning, and their transitions. It is also able to track motions in which upper and lower body are less correlated like writing on a whiteboard or boxing. The lower body pose matches the user surprisingly accurately, so the correct foot is often in contact at the right time and position. For a qualitative evaluation, we encourage the readers to compare the reconstructed simulated poses with the image references in \fref{fig:pt_new} and in the \href{https://youtu.be/CkTHsz6Ldas}{\color{blue}video}. A quantitative evaluation is shown in \tref{tab:metrics} and described in the following. 

\begin{table}[tbh]
  \caption{Pose reconstruction from synthesized and real Meta \quest HMD (H) and controllers (C). We show that despite not having sensors on the lower body, our metrics match state-of-the-art methods like PIP \cite{PIP:CVPR:2022} that use IMUs attached to the legs. Limitations are discussed in \sref{subsec:limitations}.}
  \label{tab:metrics}
  \begin{tabular}{lccccccc}
    \toprule
    
    Method                             & Ours  & Ours   & Ours & PIP \\
    Sensors                        & H+2C  & H+2C   & H  & 6 IMUs\\
    Test data                       & Lafan  & Real   & Real & TC \\
    \midrule
    MPJRE [deg]                          & 5.7     & -       & -      & -    \\
    MPJPE [cm]                           & 3.7     & -       & -        & -  \\
    RootE [cm]                            & 1.8     & -       & -       & -  \\
    SIP [deg]                          & 12.3    & -       & -        & 12.9 \\
    Jitter [km/s$^3$]                  & 0.3     & 0.1     & 0.2      & 0.2  \\
    MHPE [cm]                             & 3.7     & 6.3     & 6.2     & -  \\
    MHRE [deg]                           & 8.4     & 14.3    & 14.7    & -   \\
  \bottomrule
\end{tabular}
\end{table}

\subsection{Quantitative Evaluation}
\label{subsec:metrics}
The tracking accuracy for a variety of motions is summarized in \tref{tab:metrics}. We evaluate our trained model on synthetic input generated from the Lafan dataset \cite{2020-lafan} as well as on real data recorded while wearing the headset. We evaluate poses reconstructed using the HMD and controllers (H+2C), and using only the 6 DoF headset pose (H). We compare our metrics to a state-of-the-art solution \cite{PIP:CVPR:2022}, which uses 6 IMUs, two attached to the legs and evaluated on the TotalCapture (TC) dataset \cite{totalCapture:2017}. 

As metrics we use MPJRE (Mean Per Joint Rotation Error) in degrees and MPJPE (Mean Per Joint Position Error) in centimeters. We also compute the global root error in centimeters. To compare to \cite{PIP:CVPR:2022}, we calculate the "SIP" error, which represents the mean orientation error of the upper arms and legs in the global space in degrees. 
Jitter represents the smoothness between poses, where smaller is more smooth. This is quantified by the \textit{jerk}, the derivative of the acceleration, and calculated as done in \cite{PIP:CVPR:2022}. Finally, we compute the MHPE/MHRE (Mean Headset Position/Rotation Error), the error between the the simulated avatars head/hands (+offset) and the global position and orientation of the \quest devices.



\section{Discussion}
We discuss three applications of our framework. The main application is using sparse sensor information to reconstruct full-body poses (\sref{sec:hsac}). Then, we demonstrate how the same policy can track users of different scale (\sref{sec:res_scale}). Finally, we demonstrate how physics simulation can modify tracked motions to generate believable interactions with the environment (\sref{sec:res_sim}). We end with limitations of this approach (\sref{subsec:limitations}). 

\subsection{Headset and Controller Tracking}
\label{sec:hsac}
\begin{figure}[htb]
    \centering
    \includegraphics[width=\linewidth]{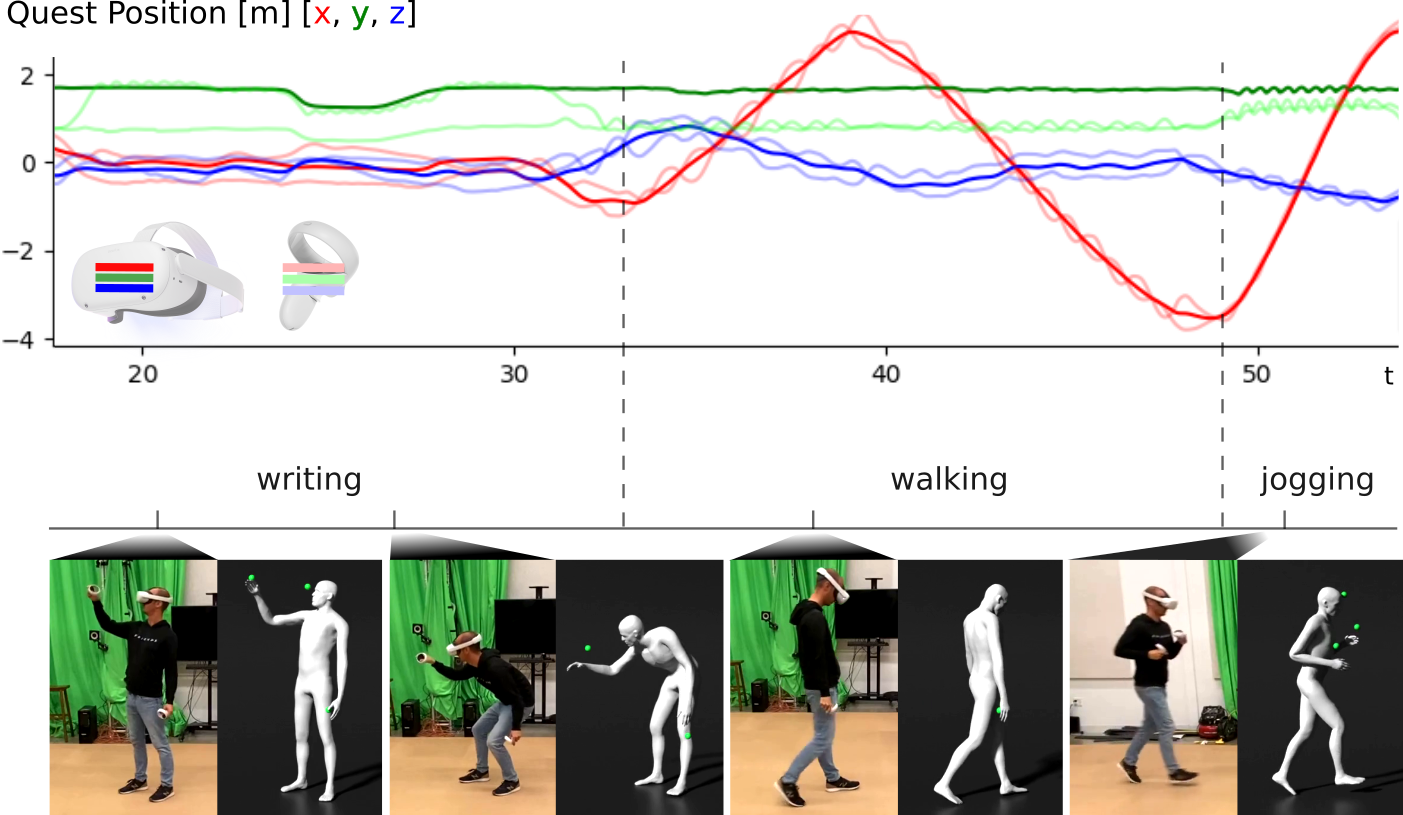}
    \caption{Sparse input used to generate the full-body pose: the position of the headset (saturated) and left and right controller (less saturated). Y is the vertical dimension. Orientation information is also used by the model but not visualized here.}
    \label{fig:plot_3pt}
\end{figure}
When comparing reconstructed poses in \fref{fig:pt_new}, we observe that the framework can distinguish different types of motion. To better understand how the model is achieving this, we plot the real sensor input driving these animations in \fref{fig:plot_3pt}. This specific clip contains three different motion types: writing, walking and jogging. Starting with the writing section, we notice that the vertical head position dips, hinting that the user is crouching and must be bending the knees. Other than that there is little correlation between controllers and head. During the walking section, the controllers oscillate in the horizontal directions (red and blue) relative to the head. This might be a pattern the model uses to identify a walking gait. In contrast to walking, jogging motions have no oscillation in the x direction (red), but displays a very unique and high frequency oscillation of the vertical HMD (green).    
%

\subsection{Tracking users of different scale}
\label{sec:res_scale}
\begin{figure}[htb]
\centering
\includegraphics[width=\linewidth]{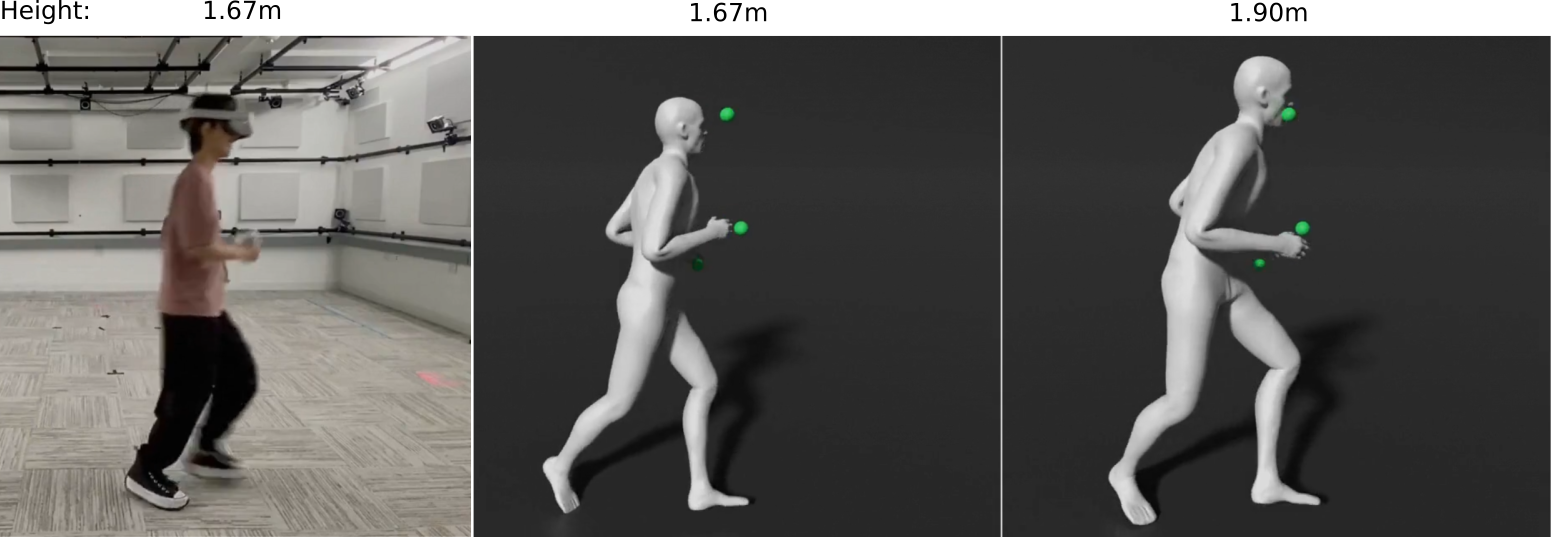}
\caption{In this example a single user controls avatars of two different sizes using the same policy.}
\label{fig:scaling}
\end{figure}
We show that a single policy can track users of different scale. The user in \fref{fig:pt_new} is 180\,cm tall, whereas the user in \fref{fig:scaling} is 167\,cm. The pose reconstruction is done by the same policy, without retraining. The user scale during inference is determined based on the initial height of the HMD, which causes the appropriately scaled avatar to be initialized (see \sref{sec:scale_obs} for details). 

Apart from tracking users of different scale, we can also use avatars of \textit{different} scale to track the \textit{same} user. Figure \ref{fig:scaling} shows the same user being tracked by a larger avatar, which is a naive form of retargeting. Since the avatar is larger, but tries to track the position of e.g. the head, it naturally results in a more crouched position than the user. But generally the avatar imitates the users motion fairly well. More importantly, due to the physics simulator, the motion doesn't violate any physical laws, such as foot skating or floor penetration that often appear when retargeting motions. With this naive baseline, future work could investigate how to further decouple users from avatar representation, allowing them to be embodied by whatever avatar they choose.

\subsection{Environment Interaction and Adaptation}
\label{sec:res_sim}
\begin{figure}[htb]
\centering
\includegraphics[width=\linewidth]{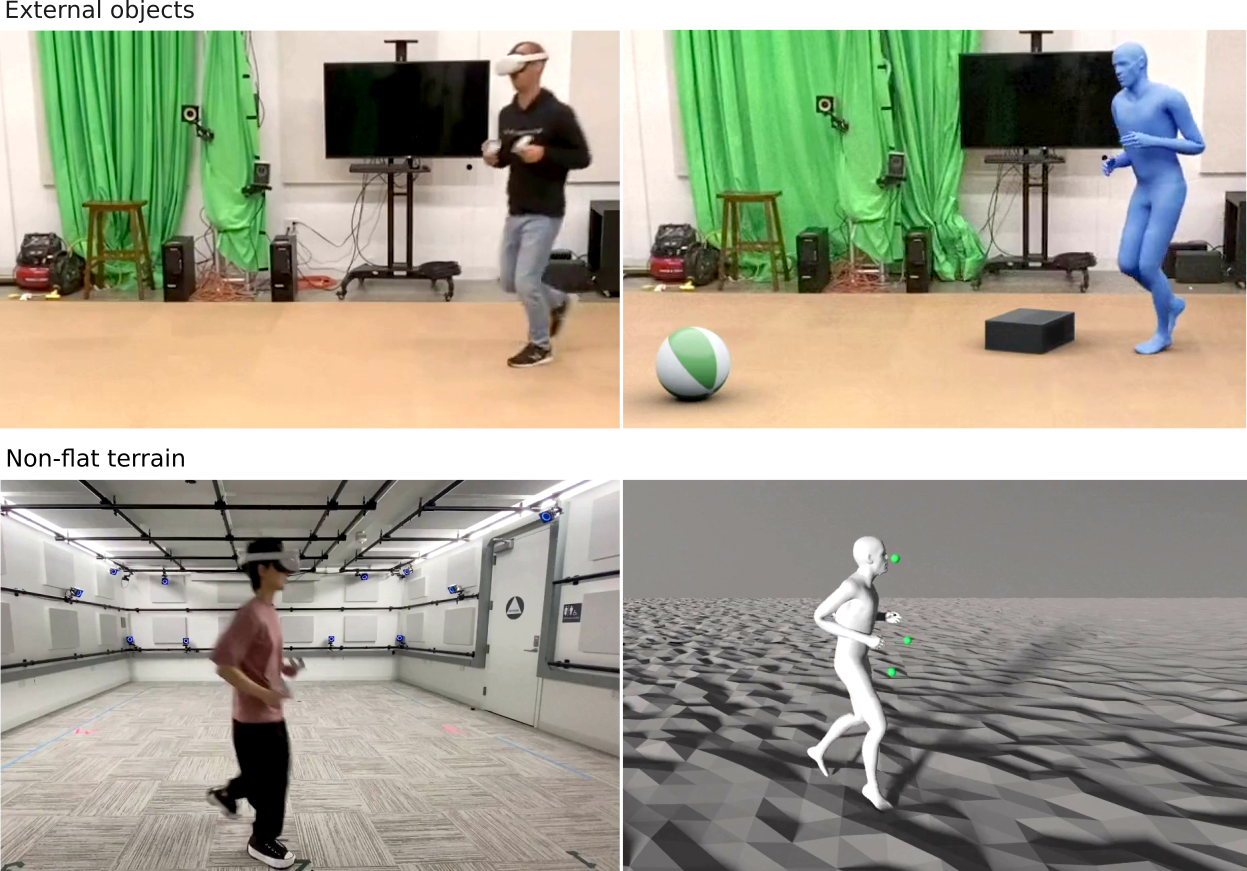}
\caption{Top: The avatar pose is influenced by the environment and it can interact with virtual objects. Bottom: If a user controls an avatar in an environment different than their play space, the motion is adapted to match the virtual world (e.g. rough ground).}
\label{fig:environment_adaptation}
\end{figure}
Since the characters are physically simulated (compared to kinematically tracked), their pose is influenced by collisions with the simulated environment (floor, objects). This adaptation of the tracked pose by physics can compensate for missing sensor data and convincingly integrate characters with their virtual world. Two scenarios are shown in Figure \ref{fig:environment_adaptation}.

In the top scenario a ball and a rigid object are loaded into the physics simulator. As a user is tracked, we observe a two-way interaction: The avatar influences the external objects (ball kick), and the objects can influence the pose of the avatar (trip hazard). This poses new research questions, for instance, how much should this semi-autonomous avatar be controlled by physics, and when should physics be violated to closely track a user?

The bottom scenario demonstrates an avatar tracking a motion on rough terrain, while the user and sensor signal is recorded on flat ground. This example demonstrates how the reconstructed motion adapts to unseen virtual worlds to create believable animations, for instance the foot adapts to different slopes in the terrain. A similar application would be a user siting on a chair, in which case the avatar's mesh should not penetrate the virtual chair. It might be difficult to measure the user to this accuracy with sensors. However, the simulated collisions and gravity can correct the tracked avatars pose to avoid penetration or hovering.


\subsection{Limitations}
\label{subsec:limitations}
While this method can produce high quality tracking results for some motions, it can also fail to track others entirely. This is different than kinematic-leaning approaches \cite{PIP:CVPR:2022}, which, on difficult motions, might perform worse (e.g more jitter, less accuracy), but nonetheless still track the motion. But since our motions are simulated without additional non-physical forces, moving the root of the character to a desired position requires a precise sequence of joint torques. Furthermore, physics simulation does not allow teleportation, so as the character drifts further from the user, it can become increasingly difficult to catch up. For these reasons the simulated character can fall when attempting to imitate a dynamic out-of-distribution motion for which it hasn't yet learned the torque controls (e.g. break-dance, jumping).  

Another difficulty comes from uncorrelated upper-lower body motions, resulting in different motions being represented by the same upper body sensor data. In this case, the policy will synthesize a natural and physically-valid lower body pose, but this might not match the user's pose.

%% file: conclusion.tex
\section{Conclusion and Future Work}
We presented a method to track users from sparse sensor data building on approaches from imitation learning. We show that physics simulation compensates for missing sensor information by synthesizing poses in a physically plausible way.

There are a variety of avenues for future work. In terms of motion quality, the simulated avatar can still look stiff and unnatural. Different reward strategies or GAN-based approaches \cite{2021-AMP-Peng} might improve the style with the goal of achieving VFX-quality animations from sparse input data. 
Secondly, due to the reliance on future observations, our real-time system currently has a latency of 160 ms. One way to reduce this and make the tracking more responsive could be to predict future poses while tracking \cite{yuan2019ego}, maybe using Motion VAEs \cite{ling2020character}.
Next, the scale value (\sref{sec:scale_obs}) is only a very coarse approximation of the user, since we linearly scale all elements (link length, mass, collision geometry, inertia) of a default skeleton equally. In reality, users of the same height can still have different proportions, resulting in different dynamics \cite{won2019learning}. In the future we want to supply the policy with more detailed skeleton and body shape information. 
Finally, we want to increase the diversity of motions the avatars can imitate. This could be achieved using mixture-of-expert policies \cite{won2020scalable, 2022-Soccer-Juggle}, pre-trained low-level controllers that facilitate learning high-level tasks \cite{2022-TOG-ASE, Jungdam2022-cVAEs}, or more informative observation representations \cite{StarkeDeepPhase2022}.

%% file: suppl_material.tex
\section{Algorithm Parameters}
\label{app:params}
The following lists the parameters used and implementation details for reproducability. An overview of the RL training procedure is given in Section 3.7. 

The policy outputs torques at a frequency of 1/36s. Both policy and value function are modeled by an MLP with 3 hidden layers of [400, 300, 200] nodes per layer and tanh activations. The Gaussian exploration noise added during training is 0.03. The output of the policy $[-1, 1]$ is scaled to $[-200, 200]$Nm. The reward uses the weights $w=[0.4, 0.1, 0.2, 0.1, 0.2]$ and Gaussian kernel sizes $k=[40.0, 0.3, 6.0, 2.0, 0.01]$. The \textit{IsaacGym} simulation frequency is 1/36s, with 2 substeps. Since the policy outputs torques, we set \textit{driveMode} in IsaacGym to DOF\_MODE\_EFFORT. We add friction of 0.1 to the joints for stability. The floor plane has a static and dynamic friction of 1.0, and a restitution of 0.0. We simulate 4000 characters in parallel, which each perform 15 steps with the same policy. This generates batch sizes of 60000, which are split into 4 minibatches. We run 5 learning epochs on each to update the policy weights $\theta$. 

We use the open-source PPO implementation in \textit{rsl\_rl} to update the policy \cite{2021-Nikita-Isaac-Code, 2021-Nikita-Isaac-Paper}. 
This approximates the gradient using Proximal Policy Optimization (PPO), with a clip parameter of 0.2. Advantages are calculated through GAE($\lambda$) \cite{2017-PPO-Schulman}. We update the Value function using targets from TD($\lambda$) \cite{sutton1998rl}. We use a discount factor $\gamma=0.97$, and $\lambda=0.95$ for the advantage estimation. The learning rate is set to $0.0001$.